\pdfoutput=1

\documentclass[11pt]{article}

\usepackage[]{acl}

\usepackage{times}
\usepackage{latexsym}
\usepackage{booktabs}
\usepackage{comment}
\usepackage{multirow}
\usepackage{tikz}
\usepackage{subcaption}  
\usepackage{linguex}
\usepackage{booktabs}
\usepackage{tabularx}
\usepackage{siunitx}
\usepackage{amsmath}
\usepackage{booktabs}
\usepackage{xcolor}

\usepackage[T1]{fontenc}

\DeclareUnicodeCharacter{2060}{\nolinebreak}
\usepackage[utf8]{inputenc}

\usepackage{inconsolata}

\usepackage{lipsum}
\usepackage{graphicx}
\usepackage{multirow}
\usepackage{array}
\usepackage{makecell}
\usepackage{tcolorbox}
\usepackage{verbatim}
\usepackage{xcolor}

\title{Leveraging Discourse Structure for Extractive Meeting Summarization}

\author{Virgile Rennard$^{1,2}$, Guokan Shang$^{3}$, Michalis Vazirgiannis$^{2,3}$, Julie Hunter$^{1}$
\\
$^1$LINAGORA, $^2$\'Ecole Polytechnique, $^3$MBZUAI
\\
\texttt{virgile@rennard.org guokan.shang@mbzuai.ac.ae}
\\
\texttt{mvazirg@lix.polytechnique.fr jhunter@linagora.com}
}

\begin{document}
\maketitle
\begin{abstract}

We introduce an extractive summarization system for meetings that leverages discourse structure to better identify salient information from complex multi-party discussions. Using discourse graphs to represent semantic relations between the contents of utterances in a meeting, we train a GNN-based node classification model to select the most important utterances, which are then combined to create an extractive summary. Experimental results on AMI and ICSI demonstrate that our approach surpasses existing text-based and graph-based extractive summarization systems, as measured by both classification and summarization metrics. Additionally, we conduct ablation studies on discourse structure and relation type to provide insights for future NLP applications leveraging discourse analysis.

\end{abstract}

\section{Introduction}

In recent years, the task of meeting summarization has garnered significant attention \cite{rennard-etal-2023-abstractive} due in part to the rising adoption of videoconferencing \cite{kost2020you} and the accumulated amount of meeting recordings, highlighting the need for automatic, novel and efficient processing methods.
Despite recent advances in natural language processing, the challenge of generating concise and coherent summaries from lengthy, multi-party meeting transcriptions remains.
Large language models suffer from the middle curse \citep{liu2024lost} and struggle to use information in the middle of their context window, particularly for summarization \citep{ravaut2023context}.

Moreover, dealing with meetings presents several challenges that are not typically encountered in traditional document summarization \citep{shang2021spoken}. Meetings are often characterized by spontaneous interaction, leading to phenomena such as poorly formed utterances and overlapping speech that negatively impact the accuracy of speech-to-text models---the first step in an automated meeting summarization pipeline. Spontaneous speech also leads to digressions, reformulations and repetitions, which can increase transcript length while simultaneously diluting information density. A further problem is the private nature of many  meetings, which greatly limits the quantity of potential public training data.

One strategy to offset data scarcity and content sparseness is to enrich conversation transcripts with additional information in the hopes of enhancing a system's dialogue understanding. One might add information about the discursive function that an utterance plays, for instance; that is, whether it serves to introduce or answer a question, to acknowledge another utterance or so on. In this vein, \citet{feng2020dialogue} showed that representing the content of a meeting transcript as a discourse graph in the style of Segmented Discourse Representation Theory \citep[SDRT;][]{asher1993reference,lascarides2008segmented} can improve performance on abstractive meeting summarization, while \citet{goo2018abstractive} demonstrated the value of dialogue acts \citep{jurafsky1997switchboard, allen1997draft} for the same task. 
Similarly, \citet{liu2019automatic} showed that supplementing transcripts with multi-modal information about participants' head orientation and eye gaze can help to identify salient information in a meeting.

In this paper, we exploit information on discourse structure to improve \textit{extractive} summarization. While abstractive summarization is generally preferable for spontaneous conversation \citep{murray-etal-2010-generating}, focusing on extractive summarization is valuable for multiple reasons.  
First, even if it is not entirely immune to hallucinations \citep{zhang-etal-2023-extractive}, extractive summarization does not suffer from this phenomenon at the level that its abstractive counterpart does \citep{cao2018faithful}. On the one hand, this makes extractive summarization an attractive final product in itself in contexts where reliability is crucial. On the other hand, this, combined with the fact that extractive summarization is easier to evaluate than abstractive summarization, makes it a clearer lens through which to study the interaction between discourse structure and content salience. Finally, as a part of a pipeline for abstractive summarization \citep{shang-etal-2020-energy}, it can be used to reduce the number of tokens given as input to a generative system with context length constraints, such as a transformer-based model.

Our approach involves a novel combination of Graph Neural Networks (GNNs) and graph-based representations of discourse in which each node in a graph represents the content of an individual utterance and each edge represents a semantic relation between two utterances whose nature is specified by a label, e.g., \textit{Explanation}, \textit{Correction}, \textit{Question-Answer Pair} or \textit{Acknowledgment}. 
The task of extractive summarization is then couched as one of binary node-classification in which nodes are determined to be important or not, and the content of the nodes judged to be important is what determines the final extractive summary.  This approach, in contrast to the generation at the whole graph-level  \citep{feng2020dialogue}, allows us to gain fine-grained insight into the interaction between the importance of an utterance in a conversation and its role in the overall discourse.
A further advantage of a graph-based approach is that GNNs, whose attention is focused only on the neighbors of a given node, do not suffer from the context length restrictions imposed by the pairwise attentional computations of transformer-based models. This means that our extractive summarization approach can take an entire long meeting transcript as input without special treatment.

To validate our discourse-structure approach, we conduct extensive experiments on the AMI \citep{mccowan2005ami} and ICSI \citep{janin2003icsi} corpora, comparing our method against diverse extractive summarization systems. The results demonstrate that our approach outperforms traditional techniques across evaluation metrics for both classification and summarization, including F1 score, ROUGE and BERTScore. Additionally, we present a detailed analysis of the impact of various graph construction strategies on summarization quality, offering insights into the mechanisms through which discourse structure influences content selection. Our study not only provides a novel methodological contribution to the field of automatic meeting summarization but also sheds light on the underlying discourse processes that shape effective summaries.
Finally, SDRT based discourse structure has seldom been explored in the literature for its potential utility in downstream tasks. Our work is among the few studies that focus on this effort, especially for dialogues.

\section{Related work}\label{sec:related}

\paragraph{Meeting summarization.} Most work focuses on abstractive summarization and tends to be organized in three interconnected streams \citep{rennard-etal-2023-abstractive}. The first focuses on enhancing transcripts with additional information---such as annotations for dialogue acts \citep{goo2018abstractive}, discourse structure \citep{feng2020dialogue} or visual cues \citep{liu2019automatic}---that is assumed to be relevant for abstracting important content from a transcript. While adding information does show improvement, few papers actually quantify how much information is gained from adding linguistic features. \citet{feng2020dialogue} used a global node to represent the discourse graph, aggregating all features into a single representative vector sent to an LSTM-based pointer decoder network for summarization; such an approach is not ideal for precisely evaluating the role of discourse structure in enhancing abstractive summaries. 

The second stream aims to transform and compress meeting transcripts in order to produce cleaner and more condensed intermediate documents that can then be passed to downstream generation modules  \citep{krishna-etal-2021-generating, oya-etal-2014-template}. The third stream is dedicated to the development of multi-task systems, such as DialogLM \citep{zhong2022dialoglm}, a language model that is specifically designed to handle dialogue content. While we focus on extractive summarization, in exploiting discourse structure, our approach overlaps with work in the first stream above and lays groundwork for  the second stream by providing a means of compressing meeting transcripts.

Turning to extractive approaches to meeting summarization, \citet{Murray2005Extractive} evaluated the effectiveness of latent semantic analysis, TF-IDF, and MMR algorithms for this task, providing benchmarks for all three methodologies. More recently, \citet{tixier-etal-2017-combining} developed a technique to construct a graph from key words in a text and then apply submodular optimization to summarize content effectively.  

We note that while significant strides have been made in developing sophisticated algorithms and techniques for meeting summarization, a problem that persists, for both extractive and abstractive summarization, is that of evaluation.  \citet{kirstein2024s} details the complications posed by different evaluation metrics and how they correlate with  various challenges specific to meeting summarization. These problems are aggravated by the lack of readily available data in English that could be used to evaluate summarization of long format dialogues \cite{janin2003icsi, mccowan2005ami, hu-etal-2023-meetingbank}---a problem that only becomes worse when we look at languages beyond English  \citep{wu2023vcsum, rennard-etal-2023-fredsum, nedoluzhko-etal-2022-elitr}.

\paragraph{Discourse structure and summarization.}
Two primary theoretical frameworks have been developed to analyze complete discourse structures in the form of graphs, namely Rhetorical Structure Theory \citep[RST;][]{mann1987rhetorical} and Segmented Discourse Representation Theory \citep[SDRT;][]{asher1993reference, lascarides2008segmented}. The framework of RST has been successfully applied for extractive summarization of documents \citep{marcu1997discourse, liu-chen-2019-exploiting, xu-etal-2020-discourse, zhu-summarizing-2021, bian2024gosum} as well as abstractive summarization \citep{pu2023incorporating}, however, these methods are not applicable to discussion by nature. However most work on SDRT discourse graphs has concentrated primarily on producing high-quality graphs \citep{shi2019deep, liu-chen-2021-improving, bennis-etal-2023-simple}, with less emphasis on their applications to downstream tasks such as summarization, and only SDRT has been applied to build discourse structures for multi-party conversation \citep{asher-etal-2016-discourse}, which is needed for meeting summarization. 

A major limitation for research on discourse structure is of course the intense effort that goes into annotating datasets for discourse structure. There are only two readily available discourse-annotated corpora for multi-party conversation: STAC \citep{asher-etal-2016-discourse}, a corpus of chats from an online version of the game \textit{Settlers of Catan}, and Molweni \citep{li-etal-2020-molweni}, an annotated version of the  Ubuntu Chat Corpus. Both corpora are annotated in the style of SDRT. 

Numerous models for building graphs have been trained on the STAC and/or Molweni corpora. \citet{shi2019deep} proposed a model that incrementally predicts both graph edges and their labels at a time $t$ by taking into account edge and label predictions before $t$. \citet{liu-chen-2021-improving} enhanced \citeauthor{shi2019deep}'s model by incorporating cross-domain information, namely by using cross-domain pretraining and vocabulary refinement. \citet{ijcai2021p543} uses a structure transformer to incorporate node and edge information of utterance pairs, and \citet{bennis-etal-2023-simple} utilizes a BERT-based framework to encode utterance pairs, predicting discourse attachments (graph edges) with a linear layer and then using a multitask approach to predict edge labels.  \citet{yang2021joint} also exploit a multi-task approach to jointly learn relation edges and labels and pronoun resolution,  demonstrating the synergistic interaction of these tasks. Finally, \citet{fan-etal-2023-improving} combined discourse parsing with addressee recognition in a multitask approach. In our evaluations, we focus on the frameworks of \citet{shi2019deep, ijcai2021p543} and \citet{bennis-etal-2023-simple} as they do not require additional data.

\section{Preliminaries}\label{sec:preliminaries}

Due to its appropriateness for multi-party dialogue \citep{asher-etal-2016-discourse} and to the fact that it is the framework adopted for the STAC and Molweni corpora, we focus on SDRT in what follows. 

\begin{figure*}[t]
\centering
\includegraphics[scale=0.4]{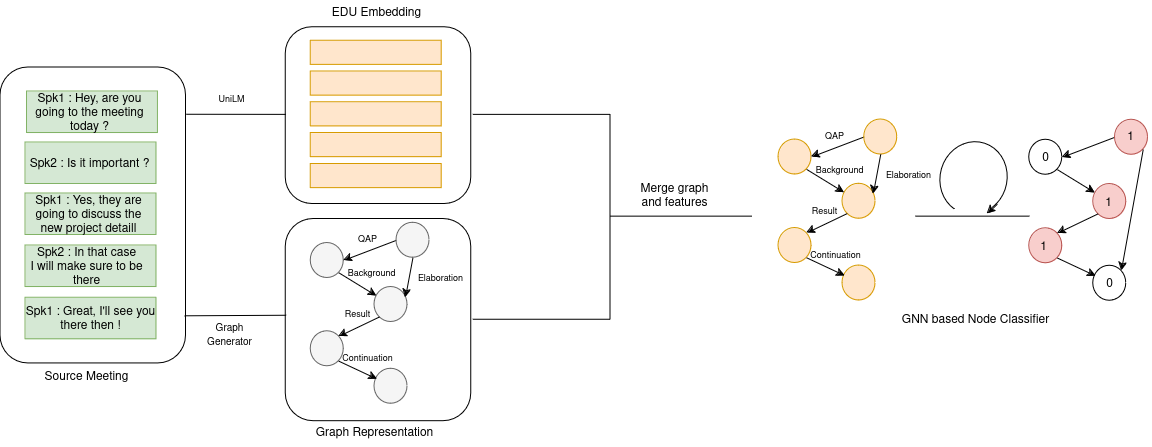}
\caption{Overview of our model. A source meeting transcript is sent to both a sentence embedding model (UniLM), which generates EDU embeddings, and to a separate discourse graph generator. The EDU embeddings are then associated with nodes of the generate graph and the resulting graph structure is sent to a GNN-based node classifier
which predicts whether an EDU should be included in a summary.}
\label{fig:model}
\end{figure*}

\subsection{Task definition}

Extractive summarization aims at creating a summary $\mathcal{Y}$ from an input meeting $\mathcal{U}$, in which $\mathcal{U}$ consists of $|\mathcal{U}|$ elementary discourse units (EDUs) $[u_1, u_2, \dots, u_{|\mathcal{U}|}]$ and $\mathcal{Y}$ consists of $|\mathcal{Y}|$ EDUs, where $|\mathcal{Y}| \leq | \mathcal{U}|$ and $[y_1,\dots,y_{|\mathcal{Y}|}] \in \mathcal{U} $. We borrow the notion of an EDU from SDRT, in which an EDU corresponds to the content of a single atomic clause so that the sentence ``The remote should glow in the dark so that it's easier to find'' would be two EDUs (the second starting with `so that')\footnote{We note that while the STAC and Molweni are segmented in this way, the transcripts of AMI and ICSI, which we use for our evaluations, are segmented slightly differently. The main difference is that when two consecutive clauses play the same discourse function in AMI and ICSI, they are not segmented. This means a sentence like ``My name is Ed and I am the project manager'' would not be segmented in these corpora because both independent clauses serve just to introduce the speaker. Because of the subtlety of this difference, we use the term EDU for simplicity.}.

It is worth noting that the input length of meeting $\mathcal{U}$ will often be longer than the input size of most language models. The average length of ICSI meetings is 13317 tokens, for instance, while  Llama's \citep{touvron2023llama} context window is 4096 tokens, which highlights the advantage of our graph-based method.

\subsection{Discourse graphs}\label{subsec:discourse_graphs}

Discourse structure in SDRT is represented as a weakly-connected graph where each node represents the content of an EDU\footnote{SDRT allows for complex discourse units (CDUs) as well, which are subgraphs, but we do not treat CDUs in this paper and they are not annotated in the corpora that we use.}, and an edge between two nodes indicates that the corresponding EDUs are related through a discourse relation. Typical SDRT relations, which are indicated as edge labels, include: \textit{Comment, Clarification-question, Elaboration, Acknowledgement, Continuation, Explanation, Conditional, Question-Answer pair, Alternation, Question-Elaboration, Result, Background, Narration, Correction, Parallel, Contrast.}

\ex.\label{ex:decision}
        \a. ID: Does anyone know if VCRs are the same across <disfmarker> international?\label{a}
        \b. PM: They're not <disfmarker> no. \label{b}
        \c. ME: They're not, no.\label{c}
        \d. ID: [Okay,] [so you'd need like a whole different set of buttons for everybody's VCRs.]\label{d}
        \e. PM: Yeah, that's right, yeah.\label{e}

\begin{figure}[ht]
  \centering
\begin{minipage}[c]{0.1\textwidth}
\begin{tikzpicture}
\node (1) at (0,-1.5) {\textbf{a}};
\node (2) at (-0.5,-2.5) {\textbf{b}};
\node (3) at (0.5,-2.5) {\textbf{c}};
\node (4) at (0,-3.5) {\textbf{d$_1$}};
\node (5) at (1.4,-3.5) {\textbf{d$_2$}};
\node (6) at (1.4,-4.5) {\textbf{e}};

\path[->,thick]
(1) edge node [left] {{\footnotesize QAP}} (2)
(1) edge node [right] {{\footnotesize QAP}} (3)
(2) edge node [left] {{\footnotesize Ack}} (4)
(3) edge node [right] {{\footnotesize Ack}} (4)
(4) edge node [below] {{\footnotesize Result}} (5)
(5) edge node [right] {{\footnotesize Ack}} (6)
;

\end{tikzpicture}
\end{minipage}\hfill
\begin{minipage}[c]{0.3\textwidth}  
  \caption{A discourse graph for example \ref{ex:decision}. Node d$_1$ represents the first EDU of \ref{d} (``Okay''), and d$_2$, the second.\\
  }
  \label{fig:discourse}
\end{minipage}
\end{figure}    

The graph for example \ref{ex:decision}\footnote{Example \ref{ex:decision} is a slightly cleaned up version of an extract from meeting ES2011b of the AMI corpus.}  is given in Figure \ref{fig:discourse}.

\section{Our model}\label{sec:model}

Our model is illustrated in Figure \ref{fig:model}. This section details the components it uses to produce an extractive summary from a  meeting transcription. 

\paragraph{EDU Embedding Module.} To perform node-classification on a graph, we first need to provide an initial representation of the nodes. Since we are dealing with text, we simply use a text embedding model to get representations for EDUs. We chose MiniLM \cite{reimers-2019-sentence-bert}, as it is a widely accessible, state-of-the-art model for sentence embeddings.

\paragraph{Graph Generator.} To generate the graph structure, we need to add edges and edge labels for different discourse relations.  Because discourse annotations require expert annotators and are very labor intensive, we opted to generate the edges and labels for our graphs with an automatic parser. In this paper, except for the ablation studies in which we consider the parsers of \citet{liu-chen-2021-improving} and \citet{bennis-etal-2023-simple}, we use Deep Sequential \cite{shi2019deep}. 

For a meeting $\mathcal{M}$ with $|N|$ EDUs, the Graph Generator outputs a meeting graph $\mathcal{M}_\mathcal{G} = (\mathbf{V}_\mathcal{M},\mathbf{E}_\mathcal{M},\mathbf{R}_\mathcal{M})$, where $\mathbf{V}_\mathcal{M}$ is the set of $|N|$ nodes $v_i$; $\mathbf{E}_\mathcal{M}$, a set of $|K|$ edges $e_{i,j} = (v_i,v_j)$ for some $K$;
and $\mathbf{R}_\mathcal{M}$, a set of $|K|$ relation labels  $r_{i,j}$ = $(v_i,v_j,R)$, where $R$ is one of the relation types listed in Subsection \ref{subsec:discourse_graphs}.

\paragraph{GNN Classifier.} With the initial features $h_i^0$ of each node $v_i \in \mathbf{V}_\mathcal{M} $ created by the EDU Embedding Module and the graph  $\mathcal{M}_\mathcal{G}$ in place, we use two different architectures to evaluate the impact of discourse structure on node classification. 

First, we use a Relational Graph Convolutional Network \citep[RGCN;][]{schlichtkrull2018modeling} to gather local hidden features weighted by the edge label. The RGCN does message passing, calculating the representation of a node $v_i$ at the step $l + 1$ ($l$ being the number of layers in the RGCN) computed as:
\begin{align}
    h_i^{(l+1)} = ReLU (\sum_{r\in \mathcal{R}_\mathcal{M}} \sum_{v_j \in \mathbf{N_i^r}} \frac{1}{|\mathbf{N}^r_i|}\mathbf{W}_r^{(l)}\mathbf{h}_j^{(l)}) \nonumber
\end{align}
in which $\mathbf{N}_i^r$ denotes the set of neighbors of node  $v_i$ under relation $r$ and $\mathbf{W}_r^{(l)}$ denotes learnable parameters at the $l$ layer for the relation $r$. Finally, we send the final representation $h_i^{l_{final}}$ to be classified by a sigmoid function on whether its content should be included in the extractive summary.

Second, to evaluate the influence of graph structure alone while ignoring edge labels, we  use MixHop GCN \citep{abu2019mixhop}. This model architecture supports multiple ``hops'',  allowing it to access and incorporate broader structural information from nodes that are more than one edge away. This relation-agnostic GNN allows us to quantify the extent to which  the graph structure itself contributes to the model's performance. 

\section{Experiments}
\label{sec:exp}

\paragraph{Datasets.} We experiment on two conversational datasets containing extractive summaries: AMI \citep{mccowan2005ami} and ICSI \citep{janin2003icsi}. Following \citet{shang-etal-2018-unsupervised}, we split AMI and ICSI into training, validation, and test sets. Basics statistics are provided in Table \ref{tab:datasets}.

\begin{table}[t]
\centering
\scalebox{0.8}{
\begin{tabular}{lcc}
    \toprule  
     & \makecell{\textbf{AMI} \\} & \makecell{\textbf{ICSI} \\}   \\
    \midrule  
    Number of meetings & 137 & 75 \\
    Average number of words & 6007.7  & 13317.3\\
    Average length of summary  & 1979.8 & 2378.3  \\
    \bottomrule 
\end{tabular}
}
\caption{Summarization corpus statistics.}
\label{tab:datasets}
\end{table}

\begin{table*}[htb]
\small
\centering
\scalebox{0.85}{%
\begin{tabular}{clccc|ccc}
    \toprule
    & & \multicolumn{3}{c}{\textbf{AMI}} & \multicolumn{3}{c}{\textbf{ICSI}} \\
    & \textbf{Model}  & \textbf{Precision} & \textbf{Recall} & \textbf{F1} & \textbf{Precision} & \textbf{Recall} & \textbf{F1} \\
    \midrule
    & Logistic Regression & \textbf{63.02} & 37.29 & 46.87 & 06.54 & 39.94 & 11.02 \\
    & MLP & 48.60 & \textbf{77.45} & 59.72 & \textbf{27.06} & 51.91 & 35.57 \\
    & GCN & 27.15 & 62.14 & 37.79 & 15.14 & 37.67 & 21.60 \\ \hline
    & RGCN & 49.51  & 73.64 & 59.25 & 23.31 & 70.66 & 35.06 \\
    & MixHop GCN  & 50.85 & 75.50 & \textbf{60.78} & 26.65 & \textbf{59.55} & \textbf{36.82} \\
    \bottomrule 
\end{tabular}%
}
\caption{Results on the AMI and ICSI datasets. Classification performance metrics shown include Precision, Recall, and F1 Score.}
\label{tab:Classification_results}
\end{table*}

\begin{table*}[t]
\small
\centering
\scalebox{0.85}{%
        \begin{tabular}{clcccc|cccc}
            \toprule
            & &  \multicolumn{4}{c}{\textbf{AMI}} & \multicolumn{4}{c}{\textbf{ICSI}} \\
            & \textbf{Model}  & \textbf{R-1} & \textbf{R-2} & \textbf{R-L} & \textbf{BERTScore} & \textbf{R-1} & \textbf{R-2} & \textbf{R-L} & \textbf{BERTScore} \\
            \midrule
            \multirow{2}{*}{ } & LEAD-N & 55.67 & 35.26 & 32.04 & 87.14 & 59.94 &  27.94 & 21.59 & 83.30  \\
            & TextRank \cite{mihalcea-tarau-2004-TextRank} & 72.92  & 53.42 & 23.94 & 83.44 & 68.19 &  40.51 & 24.21 & 84.16 \\
            & CoreRank \citep{tixier-etal-2017-combining} & 66.90 & 45.09 & 20.33 & 81.63 & 54.71 & 23.72 & 18.01 & 79.45 \\
            & BERTSumExt \cite{liu-lapata-2019-text}  & 59.54 & 35.66 & 34.14 & 85.46 & 65.92 & 35.27 & 31.09 & 84.72 \\
            & Logistic Regression & 73.20 & 58.00 & 57.85 & 88.72 & 32.05 & 19.14 & 20.19 & 84.30 \\ 
            & MLP & 77.52 & 62.12 & 59.38 & 91.94 & 69.49 & 42.26 & 37.25 & 87.20 \\
            & GCN & 74.61 & 54.32 & 50.99 & 90.42 & 67.82 & 38.98 & 34.04 & 86.51 \\
            & GPT-4 \citep{achiam2023gpt} & 66.90 & 44.93 & 43.82 & 88.11 & 62.68 & 30.18 & 27.15 & 84.26 \\ \hline
            & RGCN & 78.29 & 62.47 & 59.61 & 92.01 & 68.81 & 42.16 & 36.12 & 87.44 \\
            & MixHop GCN & 78.21 & 62.76 & \textbf{60.24} & \textbf{92.37} &69.77 & 42.84 & \textbf{37.86} & 87.95\\
            & MixHop GCN + Ranking by length  & \textbf{79.46} & \textbf{63.70} & 24.51 & \textbf{89.19} & \textbf{70.86} & \textbf{45.42} & 22.92 & \textbf{87.94} \\
            & MixHop GCN + Ranking by logits & 76.44 & 62.92 & 23.87 & 84.18 & 70.53 & 45.42 & 22.16 & 87.94 \\
            \bottomrule 
        \end{tabular}%
}
\caption{Results on the AMI and ICSI datasets with summarization performance metrics, where R-1 represents ROUGE-1, R-2 ROUGE-2, R-L ROUGE-L and BERTScore.}
\label{tab:main_results}
\end{table*}

\paragraph{Evaluation metrics.} We evaluate our model based on both node classification and content of the overall summary. For the former, we use the traditional $F_1$ score for classification. For the latter, we report the $F_1$ score for ROUGE-1, ROUGE-2 and ROUGE-L \citep{lin-2004-ROUGE}, as well as BERTScore \citep{zhang2019BERTScore} to measure the difference between a ground truth extractive summary and a system generated one. We conduct a \textbf{budgetization} process by cutting the generated summaries (from the start) to be of the average length of the ground truth summaries when evaluating our model and baselines described below.

\paragraph{Baselines.} We use a variety of models to establish baselines that are both graph-based and text-based. From the node classification perspective (see Table \ref{tab:Classification_results}), we first evaluate the strength of the input embeddings by classifying the given utterance representation with a simple \textbf{Logistical Regression}, as well as with a MultiLayer Perceptron (\textbf{MLP}). Additionally, Graph Convolutional Network \citep[\textbf{GCN};][]{kipf2016semi}, which yields a GNN baseline score that does not take any of the relations into account.

From the text summarization perspective (see Table \ref{tab:main_results}), in addition to all the above baselines we evaluate: \textbf{BERTSumExt} \citep{liu-lapata-2019-text}, which embeds individual input sentences with BERT and uses multiple inter-sentence transformer layers stacked on top of the BERT outputs to capture document-level features for extracting summaries; \textbf{TextRank} \citep{mihalcea-tarau-2004-TextRank}, a simple, graph-based and unsupervised extractive method;
\textbf{CoreRank} \citep{tixier-etal-2017-combining}, whose approach relies on a Graph-of-Words representation of the meeting, graph degeneracy and submodularity for unsupervised extractive summarization.
\textbf{GPT-4}, in which we ask to extract the most important utterances in a one-shot fashion to create the extractive summary (see Appendix \ref{app:a1}). Finally, we use \textbf{Lead-N} as a heuristic metric, which takes the first N words of the meeting as its summary, N being the budgeted length of summaries in our case.

\paragraph{Results.} We can observe from both Table \ref{tab:Classification_results} and Table \ref{tab:main_results} that our method using MixHop GCN and RGCN consistently achieves higher scores in terms of both classification and summarization metrics. Additionally, we can see that embedding-based methods such as MLP and Logistic Regression are more competitive than the heuristic-based method or unsupervised graph-based ones. We also evaluate different ranking methods for budgetization to improve the final summary. One approach involves selecting the longest utterances identified as important by the model until the required budget length is met (MixHop GCN + Ranking by length). Another approach involves selecting utterances with the highest model probability of being included in the summary until the budget is reached (MixHop GCN + Ranking by logits). Both methods enhance the summary quality. We believe that further research on ranking techniques could yield significantly better results.

\begin{table}[t]
\centering
\scalebox{0.78}{ 
\begin{tabular}{l|rrrrr|r}
\hline
\textbf{Method}  & \textbf{Gold} & \textbf{RGCN} & \textbf{Mixhop} & \textbf{LM} & \textbf{MLP} & \textbf{CP} \\ \hline
Gold             & 0             & 199           & 216         & 264         & 344         & 446  \\ \hline
RGCN             & 201           & 0             & 196         & 267         & 355         & 438  \\
Mixhop               & 184           & 204           & 0           & 257         & 351         & 392  \\
GPT-4               & 136           & 133           & 143         & 0           & 305         & -166 \\
MLP              & 56            & 45            & 49          & 95          & 0           & -1110 \\ \hline
\end{tabular}
}
\caption{Method ranking using LLM-as-a-Judge. Each row shows how many times a summary has been preferred over the model in the corresponding column. The Copeland Score is calculating by removing the number of losses to the number of wins of each models.}
\label{tab:chatgpt}
\end{table}

\subsection{Extractive summaries ranking}

Given the similarity in performance across methods, we conducted a ranking of the summaries using LLM-as-a-Judge \citep{zheng2024judging}. We rank five generated summaries based on their informativeness and fluidity using GPT-4. In our case, we evaluate the summaries for each AMI meeting by ranking the outputs of the following methods: \textbf{RGCN}, \textbf{MixHop GCN}, \textbf{GPT-4}, \textbf{MLP}, and the human-created extractive summary. To carry out the ranking, we prompt GPT-4 with: \textit{``From this meeting: \{Meeting\}, rank these five summaries in order from best to worst based on how informative and readable they are \{Summaries\}. Only output the names of the summaries in order, and nothing else.''} To ensure robustness and reduce variability, we repeat this question 20 times for each meeting.
Results in Table \ref{tab:chatgpt} show that RGCN performs best, achieving the highest Copeland Score and consistently outscoring other methods in terms of informativeness and readability, showing the importance of taking relations into account. The fact that MixHop GCN ranks second suggests that graph-based models (RGCN, MixHop GCN) produce more effective extractive summaries than GPT-4 and MLP.

\section{Ablation studies}
\label{sec:Abl}

We conducted extensive ablations to quantify the impact of both discourse structure and relation type on extractive summarization. All the experiments were run five times and reported results are the average. RGCN and MixHop GCN have three layers and 128 base parameters.

\begin{figure}[t]
\centering
\includegraphics[width=\columnwidth]{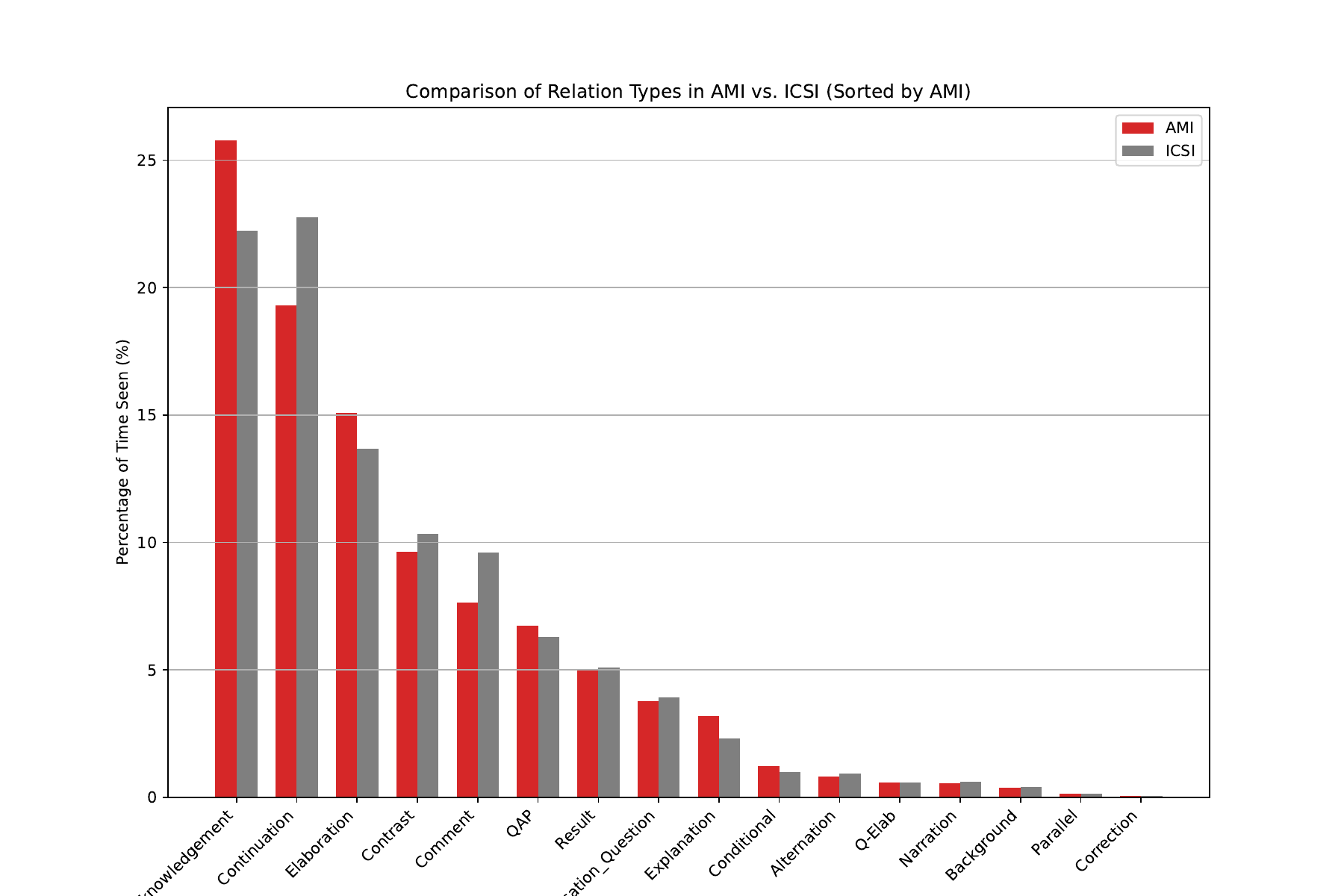}
\caption{Frequency of discourse relation types in AMI and ICSI predicted by Deep Sequential parser.}
\label{fig:frequency}
\end{figure}

\subsection{Relation type ablation}
\begin{figure*}[t]
\centering

\begin{subfigure}{0.45\textwidth}
    \centering
    \includegraphics[width=\textwidth]{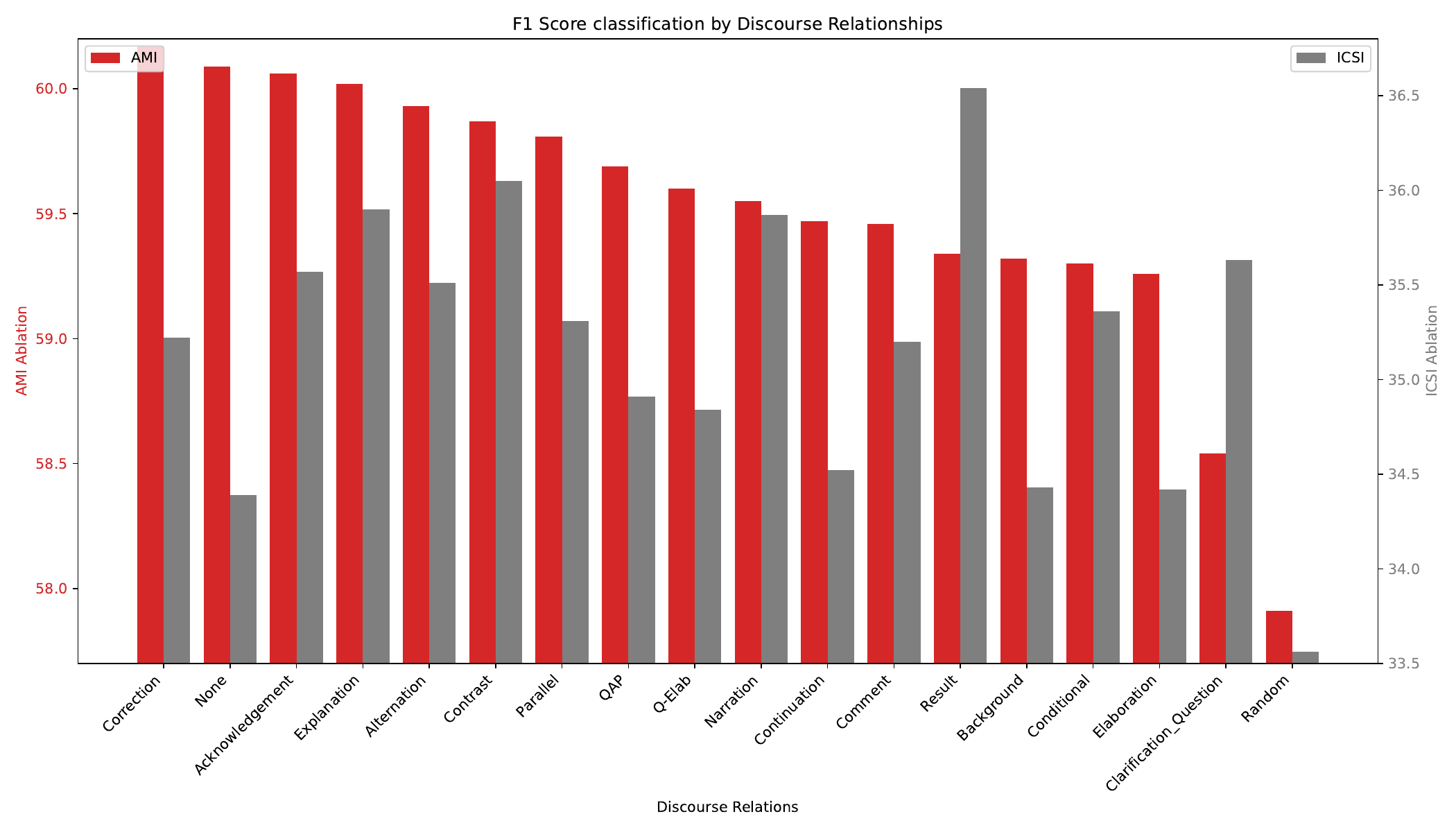}
    \caption{Impact of individual discourse relations on our RGCN classifier for both AMI and ICSI.}
    \label{fig:ablation}
\end{subfigure}
\hfill
\begin{subfigure}{0.45\textwidth}
    \centering
    \includegraphics[width=\textwidth]{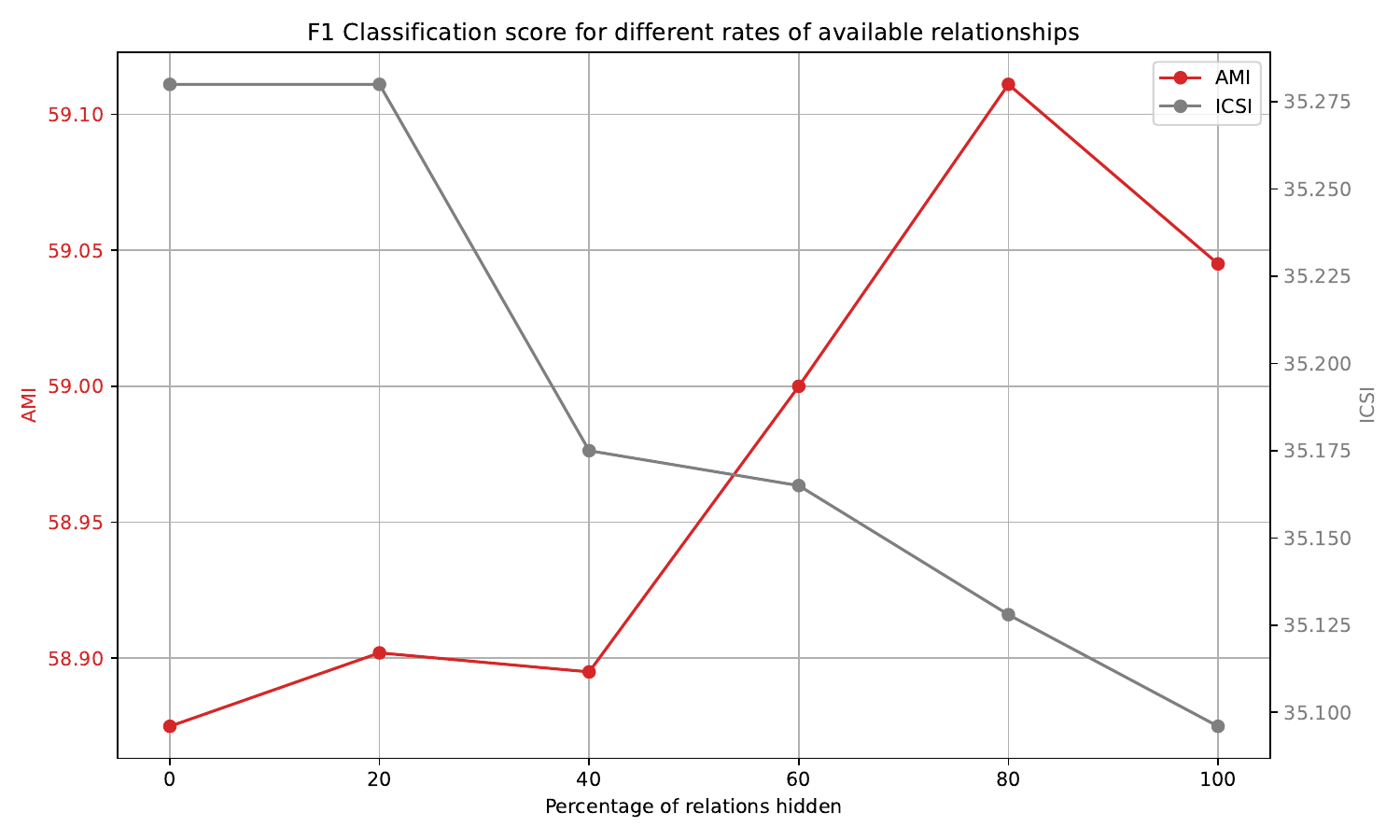}
    \caption{Impact of different rates of \textit{hidden relations} on our RGCN classifier for both AMI and ICSI.}
    \label{fig:Hidden_percent}
\end{subfigure}%

\begin{subfigure}{0.45\textwidth}
    \centering
    \includegraphics[width=\textwidth]{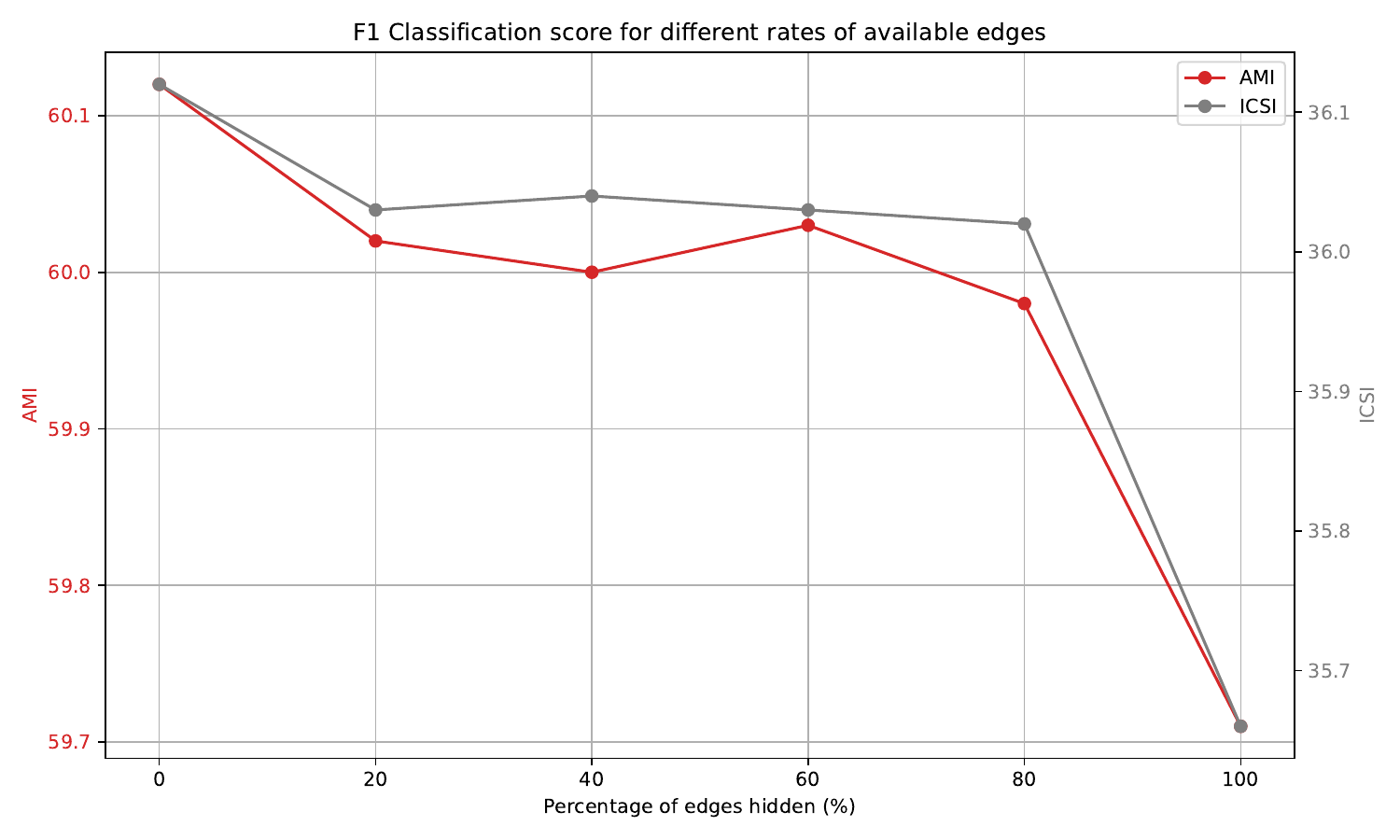}
    \caption{Impact of different rates of \textit{hidden edges} on our GCN classifier for both AMI and ICSI.}
    \label{fig:HidingEdges}
\end{subfigure}%
\hfill
\begin{subfigure}{0.45\textwidth}
    \centering
    \includegraphics[width=\textwidth]{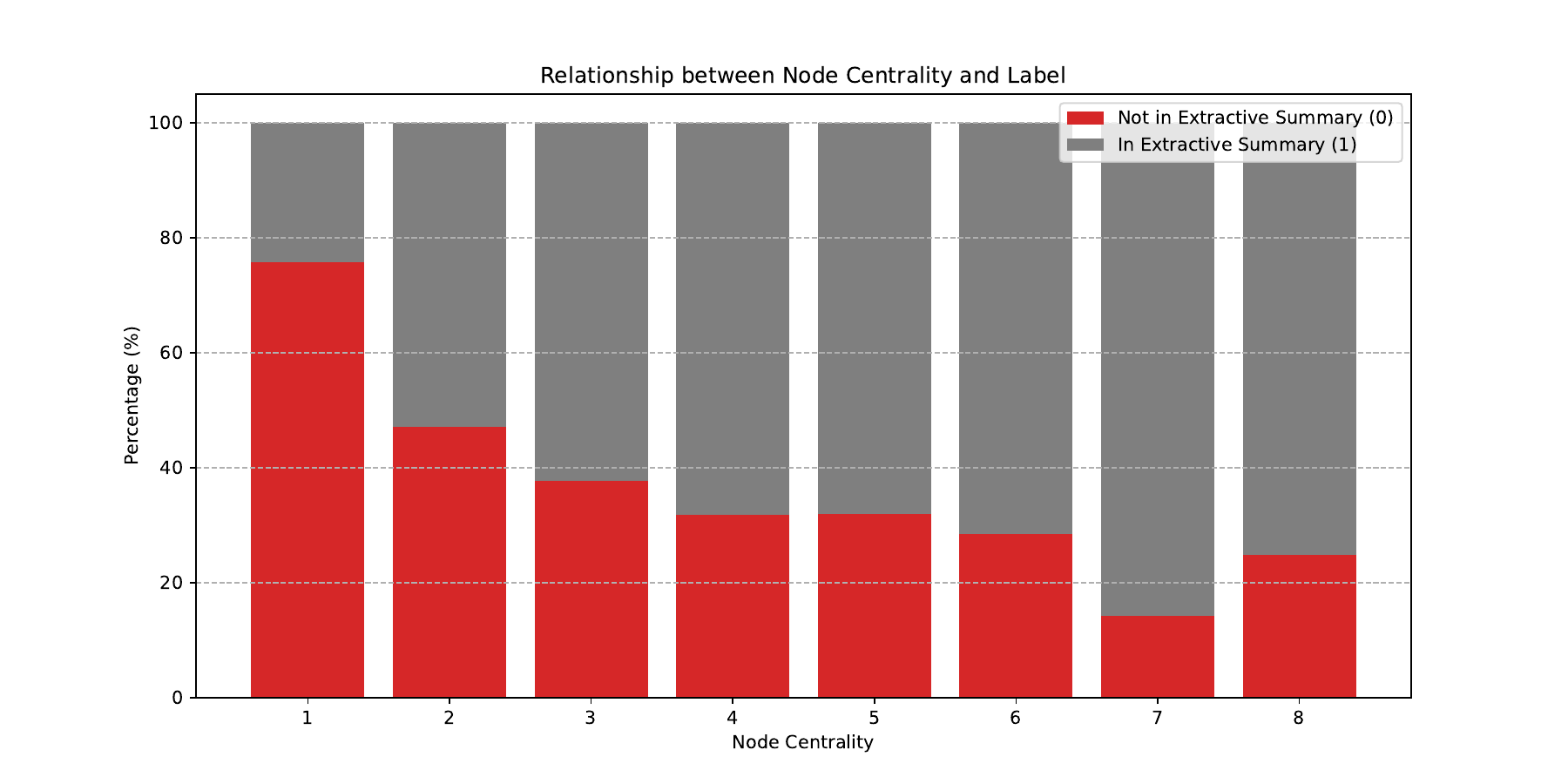}
    \caption{Percentage of node in and out of the gold summary for the different values of node centrality.}
    \label{fig:Centrality}
\end{subfigure}

\end{figure*}

As shown in Figure \ref{fig:frequency}, the Deep Sequential parser predicts a similar distribution of relations for the AMI and ICSI datasets\footnote{It is worth noting that certain relations such as \textit{Correction}, \textit{Parallel}, and \textit{Background} are extremely rare in our graphs. This rarity is due to their infrequent occurrence in the STAC corpus, on which Deep Sequential was trained.}. To evaluate the impact of each relation type on the RGCN classifier, we follow the same process as \citet{feng2020dialogue} and train an RGCN that knows only one relation type at a time. That is, we feed a graph into an RGCN for which all edges are initialized with a value of 0, with the exception of edges that correspond to the target relation. Our observations from Figure \ref{fig:ablation} indicate that within the AMI dataset,  \textit{Correction}, \textit{Acknowledgement}, and \textit{Explanation} play a significant role in classifying the EDUs, whereas for the ICSI dataset, the relations that stand out in terms of their influence on classification include \textit{Result}, \textit{Contrast}, \textit{Narration}, and \textit{Explanation}.

Additionally, Figure \ref{fig:ablation} and \ref{fig:Hidden_percent} illustrate that for the \textbf{AMI} dataset, relation labels negatively impact classification. The highest F1 scores are achieved when no relations (None) are used in the RGCN, similar to the effect of using the rarely occurring \textit{Correction} relation (see Figure \ref{fig:frequency}), which is practically equivalent to not using any relations. The RGCN also performs well when it only considers very rare relations, indicating that classification performance is driven by graph structure rather than relation types, except for \textit{Acknowledgments} (the most frequent type in Figure \ref{fig:frequency}) which, while being common, usually point to small EDUs that seldom \textit{contribute} to the summary. Conversely, for the \textbf{ICSI} dataset, the \textit{Result} relation significantly influences classification compared to other relations, establishing the importance of certain relations. Notably, in the ICSI dataset, omitting relations results in poorer classification outcomes, unlike with AMI. For both datasets, randomizing relations among nodes worsens performance (see the rightmost column in Figure \ref{fig:ablation}), indicating that nonsensical relations can hinder classification. Synergistic role of different relations are discussed in Appendix \ref{app:b}.

\subsection{Structural ablation}

We conducted further experiments to investigate the impact of graph edges on the model’s overall classification performance. As shown in Figure \ref{fig:HidingEdges},  hiding edges decreases the performance of the classifiers for both AMI and ICSI. It appears that the graph structure generally has a more robust impact on classification than relation types do. Figure \ref{fig:Centrality} shows that the more central a node is, the higher the chances of it being a part of the extractive summary is; in other words, the more common it is to refer to a certain EDU, the higher the odds of it being important. Other ways to weigh the importance of EDUs such as length of utterances, speaker changes, length of links (how many utterances between two connected nodes) do not have impact on classification.

\subsection{Studying different parsers} 

As the previously described experiments were all carried out on graphs produced by Deep Sequential \citep{shi2019deep} only, we decided to also test the impact that different graph parsers have on our model. In this section, we compare the influence of graphs produced by Deep Sequential ($DS$) with that of  graphs produced by Knowledge Enhanced \citep[$KE$;][]{liu-chen-2021-improving} and BERTLine \citep[$BL$;][]{bennis-etal-2023-simple} as well as graphs generated by GPT-4 parser we develop in the Appendix \ref{app:a2}.
Results in Table \ref{tab:3parsers} show that the four parsers lead to heterogeneous results, $KE$ scoring higher in F1 score than the other three, and $DS$ having the highest precision score. This difference suggesting we should investigate the nature of the differences between the model predictions. 
Additionally, the GPT-4 parser did not deliver a significant improvement compared to what is typically observed in other NLP tasks, suggesting that discourse parsing remains a challenge for LLMs.

\begin{table}[t]
    \centering
    \scalebox{0.9}{ 
    \begin{tabular}{@{}lccc@{}}
    \toprule
    \textbf{Model}               & \textbf{Precision} & \textbf{Recall} & \textbf{F1} \\ \midrule
    Deep Sequential        & 49.51             & 73.64          & 59.25      \\
    Knowledge Enhanced           & 48.49             & 78.75          & 59.75      \\
    BERTLine               & 47.52             & 77.34          & 58.94      \\
    GPT-4 Parser & 47.66 & 75.34 & 58.05 \\
    \bottomrule
    \end{tabular}
    }
    \caption{Performance of different discourse parsers.}
    \label{tab:3parsers}
\end{table}

    In the end, Table \ref{tab:density} shows some statistics of the graphs. Density represents a ratio between the number of edges and nodes.
    Clustering coefficient measures the degree to which vertices in a graph tend to be clustered together. It quantifies how likely it is that the neighbors of a vertex are also connected to each other. In this sense, a value of 0 shows that no two neighbors of a node are also connected (i.e., no triangle structure). We can observe that $DS$ and $KE$ graphs, have a higher number of edges but lack any triangle structure. On the other hand, we observed that $BL$ graphs have entirely disconnected nodes that will have no impact in the classification process for a GNN, explaining its relatively lower classification power.

     \begin{table}[t]
        \centering
        \label{tab:metrics2}
        \scalebox{0.75}{ 
        \begin{tabular}{@{}lcccc@{}}
        \toprule
        \textbf{Model}  & \textbf{N° Edges} & \textbf{Density} & \textbf{Cluster. Coeff.} \\ \midrule
        Deep Sequential & 755.56      & 0.003745             & 0.0          \\
        Knowledge Enhanced & 756.56 & 0.003667    & 0.0          \\
        BERTLine     & 617.40      & 0.003789             & 0.1623          \\
        GPT-4 Parser & 700.72 & 0.003344 & 0.0 \\
        \bottomrule
        \end{tabular}
        }
        \caption{Statistics of graph structures of different discourse parsers.}
        \label{tab:density}
        \end{table}

\section{Conclusion}

In this paper, we presented an extractive summarization system that leverages discourse structure to identify key information in multi-party meetings, demonstrating that discourse parsing improves summarization performance on the AMI and ICSI dataset, our method outperforming traditional text and graph-based techniques across multiple metrics, including F1 score, ROUGE, BERTScore and LLM as a judge, this ranking showing the importance of relations through its relative results on the RGCN outputs. Through ablation studies, we also analyzed the impact of relations as well as graph structure, indicating the relative role in classification for both. Our comparison of discourse parser indicating that the task is still a challenge for large language models. In conclusion, SDRT discourse parsing does hold promise for improving dialogue understanding.

\section*{Limitations}

Discourse graphs emerge as potent tools for understanding complex interactions within texts. However, as illustrated in Figure \ref{fig:frequency} a notable limitation is the under representation of certain relations within these graphs. This scarcity can be traced back to the limited availability of datasets annotated with comprehensive discourse notations, which are often specialized. Specifically, the STAC dataset is the only conversational corpus featuring gold discourse graphs. The absence of certain relations within the available data impacts the training of graph generators, tending to omit these less available relations, resulting in a generation bias towards certain structural patterns.

Moreover, while the system creates quality extractive summaries, the inherent limitations of extractive summarization for dialogues still holds. Compared to abstractive summaries, extractive summaries tend to be less fluid for readers. Addressing these readability challenges by cleaning disfluencies and refining the text remains an active area of research, underscoring the need for further advancements in dialogue summarization techniques.

Finally, as can be observed from Tables \ref{tab:Classification_results} and \ref{tab:main_results}, while our method is robustly above the others in both classification and summarization metrics, the budgetization of the summaries lowers the difference. Indeed, cutting the summaries to be the average length makes the summary focus on the early part of a meeting, and gives higher weights to false positives. Developing a strong ranking method before the budgetization would likely lead to improved results and a higher difference between classification and ROUGE scores.

\bibliography{aclanthology,googlescholar}

\begin{thebibliography}{57}
\expandafter\ifx\csname natexlab\endcsname\relax\def\natexlab#1{#1}\fi

\bibitem[{Abu-El-Haija et~al.(2019)Abu-El-Haija, Perozzi, Kapoor, Alipourfard, Lerman, Harutyunyan, Ver~Steeg, and Galstyan}]{abu2019mixhop}
Sami Abu-El-Haija, Bryan Perozzi, Amol Kapoor, Nazanin Alipourfard, Kristina Lerman, Hrayr Harutyunyan, Greg Ver~Steeg, and Aram Galstyan. 2019.
\newblock Mixhop: Higher-order graph convolutional architectures via sparsified neighborhood mixing.
\newblock In \emph{international conference on machine learning}, pages 21--29. PMLR.

\bibitem[{Achiam et~al.(2023)Achiam, Adler, Agarwal, Ahmad, Akkaya, Aleman, Almeida, Altenschmidt, Altman, Anadkat et~al.}]{achiam2023gpt}
Josh Achiam, Steven Adler, Sandhini Agarwal, Lama Ahmad, Ilge Akkaya, Florencia~Leoni Aleman, Diogo Almeida, Janko Altenschmidt, Sam Altman, Shyamal Anadkat, et~al. 2023.
\newblock Gpt-4 technical report.
\newblock \emph{arXiv preprint arXiv:2303.08774}.

\bibitem[{Allen and Core(1997)}]{allen1997draft}
James Allen and Mark Core. 1997.
\newblock Draft of {DAMSL}: Dialog act markup in several layers.

\bibitem[{Asher(1993)}]{asher1993reference}
Nicholas Asher. 1993.
\newblock Reference to abstract objects in english.

\bibitem[{Asher et~al.(2016)Asher, Hunter, Morey, Farah, and Afantenos}]{asher-etal-2016-discourse}
Nicholas Asher, Julie Hunter, Mathieu Morey, Benamara Farah, and Stergos Afantenos. 2016.
\newblock \href {https://aclanthology.org/L16-1432} {Discourse structure and dialogue acts in multiparty dialogue: the {STAC} corpus}.
\newblock In \emph{Proceedings of the Tenth International Conference on Language Resources and Evaluation ({LREC}'16)}, pages 2721--2727, Portoro{\v{z}}, Slovenia. European Language Resources Association (ELRA).

\bibitem[{Bennis et~al.(2023)Bennis, Hunter, and Asher}]{bennis-etal-2023-simple}
Zineb Bennis, Julie Hunter, and Nicholas Asher. 2023.
\newblock \href {https://aclanthology.org/2023.eacl-main.247} {A simple but effective model for attachment in discourse parsing with multi-task learning for relation labeling}.
\newblock In \emph{Proceedings of the 17th Conference of the European Chapter of the Association for Computational Linguistics}, pages 3412--3417, Dubrovnik, Croatia. Association for Computational Linguistics.

\bibitem[{Bian et~al.(2024)Bian, Huang, Zhou, Huang, and Zhu}]{bian2024gosum}
Junyi Bian, Xiaodi Huang, Hong Zhou, Tianyang Huang, and Shanfeng Zhu. 2024.
\newblock Gosum: extractive summarization of long documents by reinforcement learning and graph-organized discourse state.
\newblock \emph{Knowledge and Information Systems}, pages 1--24.

\bibitem[{Cao et~al.(2018)Cao, Wei, Li, and Li}]{cao2018faithful}
Ziqiang Cao, Furu Wei, Wenjie Li, and Sujian Li. 2018.
\newblock Faithful to the original: Fact aware neural abstractive summarization.
\newblock In \emph{Proceedings of the AAAI Conference on Artificial Intelligence}, volume~32.

\bibitem[{Fan et~al.(2023)Fan, Jiang, Li, Kong, and Zhu}]{fan-etal-2023-improving}
Yaxin Fan, Feng Jiang, Peifeng Li, Fang Kong, and Qiaoming Zhu. 2023.
\newblock \href {https://doi.org/10.18653/v1/2023.emnlp-main.526} {Improving dialogue discourse parsing via reply-to structures of addressee recognition}.
\newblock In \emph{Proceedings of the 2023 Conference on Empirical Methods in Natural Language Processing}, pages 8484--8495, Singapore. Association for Computational Linguistics.

\bibitem[{Feng et~al.(2020)Feng, Feng, Qin, and Geng}]{feng2020dialogue}
Xiachong Feng, Xiaocheng Feng, Bing Qin, and Xinwei Geng. 2020.
\newblock Dialogue discourse-aware graph model and data augmentation for meeting summarization.
\newblock \emph{In Proceeding of The 30th International Joint Conference on Artificial Intelligence}.

\bibitem[{Goo and Chen(2018)}]{goo2018abstractive}
Chih-Wen Goo and Yun-Nung Chen. 2018.
\newblock Abstractive dialogue summarization with sentence-gated modeling optimized by dialogue acts.
\newblock In \emph{2018 IEEE Spoken Language Technology Workshop (SLT)}, pages 735--742. IEEE.

\bibitem[{Hu et~al.(2023)Hu, Ganter, Deilamsalehy, Dernoncourt, Foroosh, and Liu}]{hu-etal-2023-meetingbank}
Yebowen Hu, Timothy Ganter, Hanieh Deilamsalehy, Franck Dernoncourt, Hassan Foroosh, and Fei Liu. 2023.
\newblock \href {https://doi.org/10.18653/v1/2023.acl-long.906} {{M}eeting{B}ank: A benchmark dataset for meeting summarization}.
\newblock In \emph{Proceedings of the 61st Annual Meeting of the Association for Computational Linguistics (Volume 1: Long Papers)}, pages 16409--16423, Toronto, Canada. Association for Computational Linguistics.

\bibitem[{Janin et~al.(2003)Janin, Baron, Edwards, Ellis, Gelbart, Morgan, Peskin, Pfau, Shriberg, Stolcke et~al.}]{janin2003icsi}
Adam Janin, Don Baron, Jane Edwards, Dan Ellis, David Gelbart, Nelson Morgan, Barbara Peskin, Thilo Pfau, Elizabeth Shriberg, Andreas Stolcke, et~al. 2003.
\newblock The {ICSI} meeting corpus.
\newblock In \emph{2003 IEEE International Conference on Acoustics, Speech, and Signal Processing, 2003. Proceedings.(ICASSP'03).}, volume~1, pages I--I. IEEE.

\bibitem[{Jurafsky et~al.(1997)Jurafsky, Shriberg, and Biasca}]{jurafsky1997switchboard}
Dan Jurafsky, Liz Shriberg, and Debra Biasca. 1997.
\newblock \href {https://web.stanford.edu/~jurafsky/ws97/manual.august1.html} {Switchboard {SWBD-DAMSL} shallow-discourse-function annotation coders manual}.
\newblock \emph{Institute of Cognitive Science Technical Report}.

\bibitem[{Kipf and Welling(2016)}]{kipf2016semi}
Thomas~N Kipf and Max Welling. 2016.
\newblock Semi-supervised classification with graph convolutional networks.
\newblock \emph{arXiv preprint arXiv:1609.02907}.

\bibitem[{Kirstein et~al.(2024)Kirstein, Wahle, Ruas, and Gipp}]{kirstein2024s}
Frederic Kirstein, Jan~Philip Wahle, Terry Ruas, and Bela Gipp. 2024.
\newblock What's under the hood: Investigating automatic metrics on meeting summarization.
\newblock \emph{arXiv preprint arXiv:2404.11124}.

\bibitem[{Kost(2020)}]{kost2020you}
Danielle Kost. 2020.
\newblock You’re right! {Y}ou are working longer and attending more meetings.
\newblock \emph{Harvard Business School Working Knowledge}.

\bibitem[{Krishna et~al.(2021)Krishna, Khosla, Bigham, and Lipton}]{krishna-etal-2021-generating}
Kundan Krishna, Sopan Khosla, Jeffrey Bigham, and Zachary~C. Lipton. 2021.
\newblock \href {https://doi.org/10.18653/v1/2021.acl-long.384} {Generating {SOAP} notes from doctor-patient conversations using modular summarization techniques}.
\newblock In \emph{Proceedings of the 59th Annual Meeting of the Association for Computational Linguistics and the 11th International Joint Conference on Natural Language Processing (Volume 1: Long Papers)}, pages 4958--4972, Online. Association for Computational Linguistics.

\bibitem[{Lascarides and Asher(2008)}]{lascarides2008segmented}
Alex Lascarides and Nicholas Asher. 2008.
\newblock Segmented discourse representation theory: Dynamic semantics with discourse structure.
\newblock In \emph{Computing meaning}, pages 87--124. Springer.

\bibitem[{Li et~al.(2020)Li, Liu, Kan, Zheng, Wang, Lei, Liu, and Qin}]{li-etal-2020-molweni}
Jiaqi Li, Ming Liu, Min-Yen Kan, Zihao Zheng, Zekun Wang, Wenqiang Lei, Ting Liu, and Bing Qin. 2020.
\newblock \href {https://doi.org/10.18653/v1/2020.coling-main.238} {Molweni: A challenge multiparty dialogues-based machine reading comprehension dataset with discourse structure}.
\newblock In \emph{Proceedings of the 28th International Conference on Computational Linguistics}, pages 2642--2652, Barcelona, Spain (Online). International Committee on Computational Linguistics.

\bibitem[{Lin(2004)}]{lin-2004-ROUGE}
Chin-Yew Lin. 2004.
\newblock \href {https://aclanthology.org/W04-1013} {{ROUGE}: A package for automatic evaluation of summaries}.
\newblock In \emph{Text Summarization Branches Out}, pages 74--81, Barcelona, Spain. Association for Computational Linguistics.

\bibitem[{Liu et~al.(2019)Liu, Wang, Xu, Li, and Ye}]{liu2019automatic}
Chunyi Liu, Peng Wang, Jiang Xu, Zang Li, and Jieping Ye. 2019.
\newblock Automatic dialogue summary generation for customer service.
\newblock In \emph{Proceedings of the 25th ACM SIGKDD International Conference on Knowledge Discovery \& Data Mining}, pages 1957--1965.

\bibitem[{Liu et~al.(2024)Liu, Lin, Hewitt, Paranjape, Bevilacqua, Petroni, and Liang}]{liu2024lost}
Nelson~F Liu, Kevin Lin, John Hewitt, Ashwin Paranjape, Michele Bevilacqua, Fabio Petroni, and Percy Liang. 2024.
\newblock Lost in the middle: How language models use long contexts.
\newblock \emph{Transactions of the Association for Computational Linguistics}, 12:157--173.

\bibitem[{Liu and Lapata(2019)}]{liu-lapata-2019-text}
Yang Liu and Mirella Lapata. 2019.
\newblock \href {https://doi.org/10.18653/v1/D19-1387} {Text summarization with pretrained encoders}.
\newblock In \emph{Proceedings of the 2019 Conference on Empirical Methods in Natural Language Processing and the 9th International Joint Conference on Natural Language Processing (EMNLP-IJCNLP)}, pages 3730--3740, Hong Kong, China. Association for Computational Linguistics.

\bibitem[{Liu and Chen(2019)}]{liu-chen-2019-exploiting}
Zhengyuan Liu and Nancy Chen. 2019.
\newblock \href {https://doi.org/10.18653/v1/D19-5415} {Exploiting discourse-level segmentation for extractive summarization}.
\newblock In \emph{Proceedings of the 2nd Workshop on New Frontiers in Summarization}, pages 116--121, Hong Kong, China. Association for Computational Linguistics.

\bibitem[{Liu and Chen(2021)}]{liu-chen-2021-improving}
Zhengyuan Liu and Nancy Chen. 2021.
\newblock \href {https://doi.org/10.18653/v1/2021.codi-main.11} {Improving multi-party dialogue discourse parsing via domain integration}.
\newblock In \emph{Proceedings of the 2nd Workshop on Computational Approaches to Discourse}, pages 122--127, Punta Cana, Dominican Republic and Online. Association for Computational Linguistics.

\bibitem[{Maekawa et~al.(2024)Maekawa, Hirao, Kamigaito, and Okumura}]{maekawa-etal-2024-obtain}
Aru Maekawa, Tsutomu Hirao, Hidetaka Kamigaito, and Manabu Okumura. 2024.
\newblock \href {https://aclanthology.org/2024.eacl-long.171} {Can we obtain significant success in {RST} discourse parsing by using large language models?}
\newblock In \emph{Proceedings of the 18th Conference of the European Chapter of the Association for Computational Linguistics (Volume 1: Long Papers)}, pages 2803--2815, St. Julian{'}s, Malta. Association for Computational Linguistics.

\bibitem[{Mann and Thompson(1987)}]{mann1987rhetorical}
William~C. Mann and Sandra~A. Thompson. 1987.
\newblock Rhetorical structure theory: A framework for the analysis of texts.
\newblock Technical report, University of Southern California Marina Del Rey Information Sciences Inst.

\bibitem[{Marcu(1997)}]{marcu1997discourse}
Daniel Marcu. 1997.
\newblock From discourse structures to text summaries.
\newblock In \emph{Intelligent Scalable Text Summarization}.

\bibitem[{Mccowan et~al.(2005)Mccowan, Carletta, Kraaij, Ashby, Bourban, Flynn, Guillemot, Hain, Kadlec, Karaiskos, Kronenthal, Lathoud, Lincoln, Lisowska~Masson, Post, Reidsma, and Wellner}]{mccowan2005ami}
Iain Mccowan, Jean Carletta, Wessel Kraaij, Simone Ashby, Sebastien Bourban, Mike Flynn, Ma{\"e}l Guillemot, Thomas Hain, Jaroslav Kadlec, Vasilis Karaiskos, Melissa Kronenthal, Guillaume Lathoud, Mike Lincoln, Agnes Lisowska~Masson, Wilfried Post, Dennis Reidsma, and P.~Wellner. 2005.
\newblock The {AMI} meeting corpus.
\newblock \emph{Int'l. Conf. on Methods and Techniques in Behavioral Research}.

\bibitem[{Mihalcea and Tarau(2004)}]{mihalcea-tarau-2004-TextRank}
Rada Mihalcea and Paul Tarau. 2004.
\newblock \href {https://aclanthology.org/W04-3252} {{T}ext{R}ank: Bringing order into text}.
\newblock In \emph{Proceedings of the 2004 Conference on Empirical Methods in Natural Language Processing}, pages 404--411, Barcelona, Spain. Association for Computational Linguistics.

\bibitem[{Murray et~al.(2010)Murray, Carenini, and Ng}]{murray-etal-2010-generating}
Gabriel Murray, Giuseppe Carenini, and Raymond Ng. 2010.
\newblock \href {https://aclanthology.org/W10-4211} {Generating and validating abstracts of meeting conversations: a user study}.
\newblock In \emph{Proceedings of the 6th International Natural Language Generation Conference}. Association for Computational Linguistics.

\bibitem[{Murray et~al.(2005)Murray, Renals, and Carletta}]{Murray2005Extractive}
Gabriel Murray, Steve Renals, and Jean Carletta. 2005.
\newblock \href {https://api.semanticscholar.org/CorpusID:5776046} {Extractive summarization of meeting recordings}.
\newblock In \emph{Interspeech}.

\bibitem[{Nedoluzhko et~al.(2022)Nedoluzhko, Singh, Hled{\'\i}kov{\'a}, Ghosal, and Bojar}]{nedoluzhko-etal-2022-elitr}
Anna Nedoluzhko, Muskaan Singh, Marie Hled{\'\i}kov{\'a}, Tirthankar Ghosal, and Ond{\v{r}}ej Bojar. 2022.
\newblock \href {https://aclanthology.org/2022.lrec-1.340} {{ELITR} minuting corpus: A novel dataset for automatic minuting from multi-party meetings in {E}nglish and {C}zech}.
\newblock In \emph{Proceedings of the Thirteenth Language Resources and Evaluation Conference}, pages 3174--3182, Marseille, France. European Language Resources Association.

\bibitem[{Oya et~al.(2014)Oya, Mehdad, Carenini, and Ng}]{oya-etal-2014-template}
Tatsuro Oya, Yashar Mehdad, Giuseppe Carenini, and Raymond Ng. 2014.
\newblock \href {https://doi.org/10.3115/v1/W14-4407} {A template-based abstractive meeting summarization: Leveraging summary and source text relationships}.
\newblock In \emph{Proceedings of the 8th International Natural Language Generation Conference ({INLG})}, pages 45--53, Philadelphia, Pennsylvania, U.S.A. Association for Computational Linguistics.

\bibitem[{Pu et~al.(2023)Pu, Wang, and Demberg}]{pu2023incorporating}
Dongqi Pu, Yifan Wang, and Vera Demberg. 2023.
\newblock Incorporating distributions of discourse structure for long document abstractive summarization.
\newblock \emph{arXiv preprint arXiv:2305.16784}.

\bibitem[{Ravaut et~al.(2023)Ravaut, Joty, Sun, and Chen}]{ravaut2023context}
Mathieu Ravaut, Shafiq Joty, Aixin Sun, and Nancy~F Chen. 2023.
\newblock On context utilization in summarization with large language models.
\newblock \emph{arXiv e-prints}, pages arXiv--2310.

\bibitem[{Reimers and Gurevych(2019)}]{reimers-2019-sentence-bert}
Nils Reimers and Iryna Gurevych. 2019.
\newblock \href {https://arxiv.org/abs/1908.10084} {Sentence-bert: Sentence embeddings using siamese bert-networks}.
\newblock In \emph{Proceedings of the 2019 Conference on Empirical Methods in Natural Language Processing}. Association for Computational Linguistics.

\bibitem[{Rennard et~al.(2023{\natexlab{a}})Rennard, Shang, Grari, Hunter, and Vazirgiannis}]{rennard-etal-2023-fredsum}
Virgile Rennard, Guokan Shang, Damien Grari, Julie Hunter, and Michalis Vazirgiannis. 2023{\natexlab{a}}.
\newblock \href {https://doi.org/10.18653/v1/2023.findings-emnlp.280} {{FREDS}um: A dialogue summarization corpus for {F}rench political debates}.
\newblock In \emph{Findings of the Association for Computational Linguistics: EMNLP 2023}, pages 4241--4253, Singapore. Association for Computational Linguistics.

\bibitem[{Rennard et~al.(2023{\natexlab{b}})Rennard, Shang, Hunter, and Vazirgiannis}]{rennard-etal-2023-abstractive}
Virgile Rennard, Guokan Shang, Julie Hunter, and Michalis Vazirgiannis. 2023{\natexlab{b}}.
\newblock \href {https://doi.org/10.1162/tacl_a_00578} {Abstractive meeting summarization: A survey}.
\newblock \emph{Transactions of the Association for Computational Linguistics}, 11:861--884.

\bibitem[{Schlichtkrull et~al.(2018)Schlichtkrull, Kipf, Bloem, Berg, Titov, and Welling}]{schlichtkrull2018modeling}
Michael Schlichtkrull, Thomas~N. Kipf, Peter Bloem, Rianne van~den Berg, Ivan Titov, and Max Welling. 2018.
\newblock Modeling relational data with graph convolutional networks.
\newblock In \emph{European semantic web conference}, pages 593--607. Springer.

\bibitem[{Shang(2021)}]{shang2021spoken}
Guokan Shang. 2021.
\newblock \emph{Spoken Language Understanding for Abstractive Meeting Summarization}.
\newblock Ph.D. thesis, Institut Polytechnique de Paris.

\bibitem[{Shang et~al.(2018)Shang, Ding, Zhang, Tixier, Meladianos, Vazirgiannis, and Lorr{\'e}}]{shang-etal-2018-unsupervised}
Guokan Shang, Wensi Ding, Zekun Zhang, Antoine Tixier, Polykarpos Meladianos, Michalis Vazirgiannis, and Jean-Pierre Lorr{\'e}. 2018.
\newblock \href {https://doi.org/10.18653/v1/P18-1062} {Unsupervised abstractive meeting summarization with multi-sentence compression and budgeted submodular maximization}.
\newblock In \emph{Proceedings of the 56th Annual Meeting of the Association for Computational Linguistics (Volume 1: Long Papers)}, pages 664--674, Melbourne, Australia. Association for Computational Linguistics.

\bibitem[{Shang et~al.(2020)Shang, Tixier, Vazirgiannis, and Lorr{\'e}}]{shang-etal-2020-energy}
Guokan Shang, Antoine Tixier, Michalis Vazirgiannis, and Jean-Pierre Lorr{\'e}. 2020.
\newblock \href {https://aclanthology.org/2020.aacl-main.34} {Energy-based self-attentive learning of abstractive communities for spoken language understanding}.
\newblock In \emph{Proceedings of the 1st Conference of the Asia-Pacific Chapter of the Association for Computational Linguistics and the 10th International Joint Conference on Natural Language Processing}, pages 313--327, Suzhou, China. Association for Computational Linguistics.

\bibitem[{Shi and Huang(2019)}]{shi2019deep}
Zhouxing Shi and Minlie Huang. 2019.
\newblock A deep sequential model for discourse parsing on multi-party dialogues.
\newblock In \emph{Proceedings of the AAAI Conference on Artificial Intelligence}, volume~33, pages 7007--7014.

\bibitem[{Thompson et~al.(2024)Thompson, Chaturvedi, Hunter, and Asher}]{thompson2024llamipa}
Kate Thompson, Akshay Chaturvedi, Julie Hunter, and Nicholas Asher. 2024.
\newblock Llamipa: An incremental discourse parser.
\newblock \emph{arXiv preprint arXiv:2406.18256}.

\bibitem[{Tixier et~al.(2017)Tixier, Meladianos, and Vazirgiannis}]{tixier-etal-2017-combining}
Antoine Tixier, Polykarpos Meladianos, and Michalis Vazirgiannis. 2017.
\newblock \href {https://doi.org/10.18653/v1/W17-4507} {Combining graph degeneracy and submodularity for unsupervised extractive summarization}.
\newblock In \emph{Proceedings of the Workshop on New Frontiers in Summarization}, pages 48--58, Copenhagen, Denmark. Association for Computational Linguistics.

\bibitem[{Touvron et~al.(2023)Touvron, Martin, Stone, Albert, Almahairi, Babaei, Bashlykov, Batra, Bhargava, Bhosale et~al.}]{touvron2023llama}
Hugo Touvron, Louis Martin, Kevin Stone, Peter Albert, Amjad Almahairi, Yasmine Babaei, Nikolay Bashlykov, Soumya Batra, Prajjwal Bhargava, Shruti Bhosale, et~al. 2023.
\newblock Llama 2: Open foundation and fine-tuned chat models.
\newblock \emph{arXiv preprint arXiv:2307.09288}.

\bibitem[{Wang et~al.(2021)Wang, Song, Jiang, Lai, Yao, Zhang, and Su}]{ijcai2021p543}
Ante Wang, Linfeng Song, Hui Jiang, Shaopeng Lai, Junfeng Yao, Min Zhang, and Jinsong Su. 2021.
\newblock \href {https://doi.org/10.24963/ijcai.2021/543} {A structure self-aware model for discourse parsing on multi-party dialogues}.
\newblock In \emph{Proceedings of the Thirtieth International Joint Conference on Artificial Intelligence, {IJCAI-21}}, pages 3943--3949. International Joint Conferences on Artificial Intelligence Organization.
\newblock Main Track.

\bibitem[{Wu et~al.(2023)Wu, Zhan, Tan, Hou, Liang, and Song}]{wu2023vcsum}
Han Wu, Mingjie Zhan, Haochen Tan, Zhaohui Hou, Ding Liang, and Linqi Song. 2023.
\newblock Vcsum: A versatile chinese meeting summarization dataset.
\newblock \emph{arXiv preprint arXiv:2305.05280}.

\bibitem[{Xu et~al.(2020)Xu, Gan, Cheng, and Liu}]{xu-etal-2020-discourse}
Jiacheng Xu, Zhe Gan, Yu~Cheng, and Jingjing Liu. 2020.
\newblock \href {https://doi.org/10.18653/v1/2020.acl-main.451} {Discourse-aware neural extractive text summarization}.
\newblock In \emph{Proceedings of the 58th Annual Meeting of the Association for Computational Linguistics}, pages 5021--5031, Online. Association for Computational Linguistics.

\bibitem[{Yang et~al.(2021)Yang, Xu, Xu, Li, Gao, Guo, Xue, and Wen}]{yang2021joint}
Jingxuan Yang, Kerui Xu, Jun Xu, Si~Li, Sheng Gao, Jun Guo, Nianwen Xue, and Ji-Rong Wen. 2021.
\newblock A joint model for dropped pronoun recovery and conversational discourse parsing in chinese conversational speech.
\newblock \emph{arXiv preprint arXiv:2106.03345}.

\bibitem[{Zhang et~al.(2023)Zhang, Wan, and Bansal}]{zhang-etal-2023-extractive}
Shiyue Zhang, David Wan, and Mohit Bansal. 2023.
\newblock \href {https://doi.org/10.18653/v1/2023.acl-long.120} {Extractive is not faithful: An investigation of broad unfaithfulness problems in extractive summarization}.
\newblock In \emph{Proceedings of the 61st Annual Meeting of the Association for Computational Linguistics (Volume 1: Long Papers)}, pages 2153--2174, Toronto, Canada. Association for Computational Linguistics.

\bibitem[{Zhang et~al.(2020)Zhang, Kishore, Wu, Weinberger, and Artzi}]{zhang2019BERTScore}
Tianyi Zhang, Varsha Kishore, Felix Wu, Kilian~Q. Weinberger, and Yoav Artzi. 2020.
\newblock \href {https://openreview.net/forum?id=SkeHuCVFDr} {{BERTScore}: Evaluating text generation with {BERT}}.
\newblock In \emph{8th International Conference on Learning Representations, {ICLR} 2020, Addis Ababa, Ethiopia, April 26-30, 2020}. OpenReview.net.

\bibitem[{Zheng et~al.(2024)Zheng, Chiang, Sheng, Zhuang, Wu, Zhuang, Lin, Li, Li, Xing et~al.}]{zheng2024judging}
Lianmin Zheng, Wei-Lin Chiang, Ying Sheng, Siyuan Zhuang, Zhanghao Wu, Yonghao Zhuang, Zi~Lin, Zhuohan Li, Dacheng Li, Eric Xing, et~al. 2024.
\newblock Judging llm-as-a-judge with mt-bench and chatbot arena.
\newblock \emph{Advances in Neural Information Processing Systems}, 36.

\bibitem[{Zhong et~al.(2022)Zhong, Liu, Xu, Zhu, and Zeng}]{zhong2022dialoglm}
Ming Zhong, Yang Liu, Yichong Xu, Chenguang Zhu, and Michael Zeng. 2022.
\newblock Dialog{LM}: Pre-trained model for long dialogue understanding and summarization.
\newblock In \emph{Proceedings of the AAAI Conference on Artificial Intelligence}, volume~36, pages 11765--11773.

\bibitem[{Zhu et~al.(2021)Zhu, Hua, Qu, and Zhou}]{zhu-summarizing-2021}
Tianyu Zhu, Wen Hua, Jianfeng Qu, and Xiaofang Zhou. 2021.
\newblock \href {https://doi.org/10.1145/3459637.3482396} {Summarizing long-form document with rich discourse information}.
\newblock In \emph{Proceedings of the 30th ACM International Conference on Information \& Knowledge Management}, CIKM '21, page 2770–2779, New York, NY, USA. Association for Computing Machinery.

\end{thebibliography}

\appendix

\section{GPT-4 for extractive summarization and graph generation}
\label{sec:appendixGPT}

\begin{figure*}[t]
\centering

\begin{subfigure}{0.45\textwidth}
    \centering
    \includegraphics[width=\textwidth]{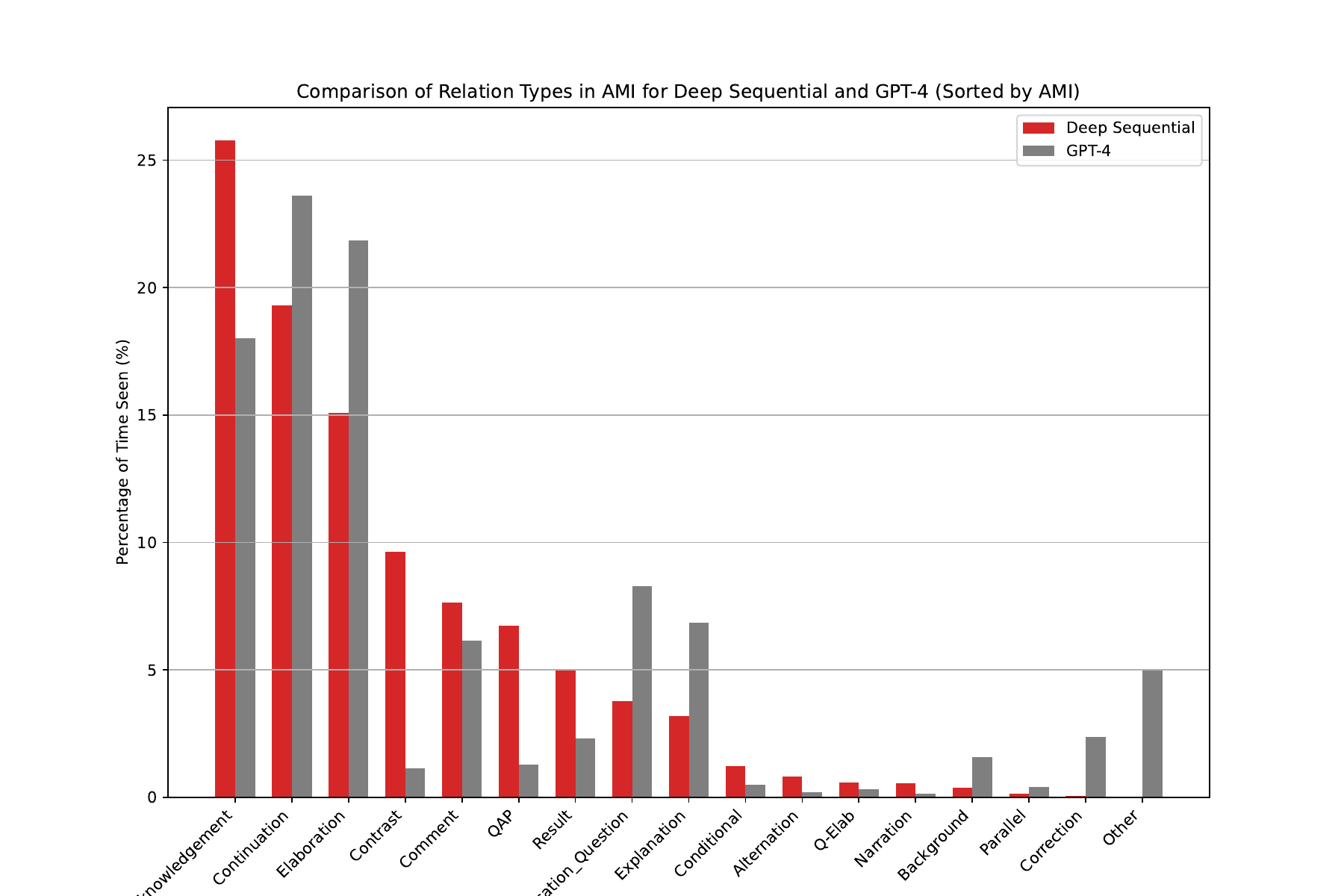}
    \caption{Comparison of the appearance of different discourse relationships for AMI graphs generated by both Deep Sequential parser and GPT-4.}
    \label{fig:spreadGPT}
\end{subfigure}%
\hfill
\begin{subfigure}{0.45\textwidth}
    \centering
    \includegraphics[width=\textwidth]{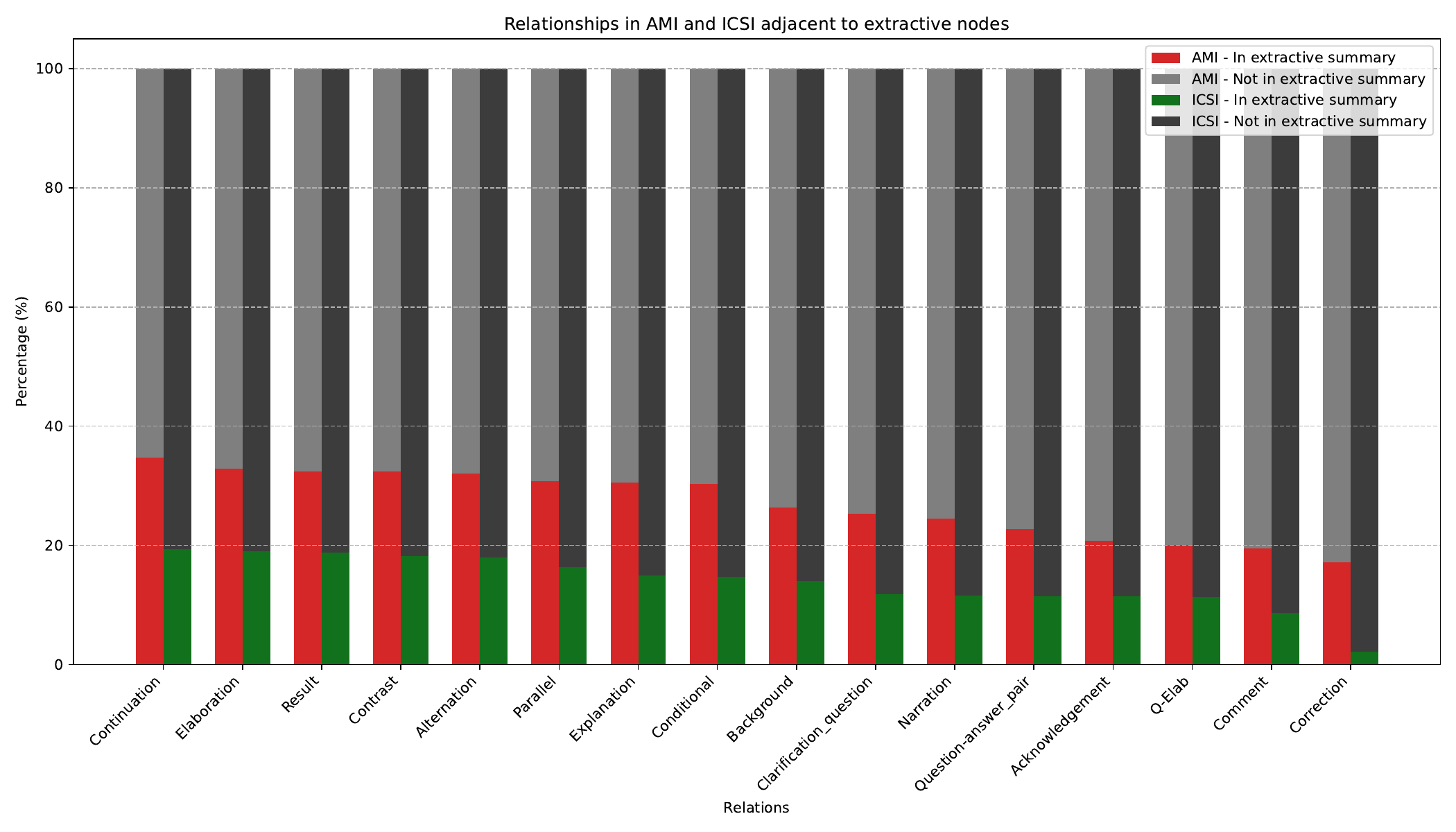}
    \caption{Comparison of relations adjacent to extractive nodes in AMI and ICSI.}
    \label{fig:frequencyGPT}
\end{subfigure}

\caption{Comparison of relations adjacent to extractive nodes and different discourse relationships in AMI and ICSI datasets.}

\end{figure*}

In Section \ref{sec:exp} and \ref{sec:Abl}, we propose using GPT-4 for both extractive summarization and discourse parsing.
In this appendix, we explain the methods used, as these tasks are not commonly performed with the help of LLMs.

\subsection{Extractive summarization} \label{app:a1}

While language models like GPT-4 excel in generative tasks, using them for discriminative tasks like extractive summarization presents specific challenges.
Asking a language model to produce an extractive summary is not straightforward, as it often generates a bullet-point summary of the meeting or rewrites disfluent text for better readability, complicating the evaluation of extraction performance.	

In our experiments, we supplied GPT-4 with a formatted version of the dialogue acts from the AMI transcripts.	
The transcripts were structured as follows:	

\begin{verbatim}
"IS1003d" : {
    "0": "Mm-hmm",
    "1": "So, ready?",
    "2": "No not really",
    "3": "'Kay",
    "4": "Just <disfmarker>",
    "5": "Crap.",
    "6": "<vocalsound> Sorry,",
    ...,
    "1091": "Very productive.",
    "1092": "<vocalsound> Okay.",
    "1093": "Thanks.",
      "1094": "S so who is going to take it?"
}
\end{verbatim}

We instructed GPT-4 to extract both the important utterances with their corresponding indices, and to also provide the indices alone.
Despite clear instructions, the model frequently omitted details like \texttt{<vocalsound>} tags or disfluencies.	
To simplify our evaluations, we eventually asked GPT-4 to return only the indices of important utterances.	
This enabled us to apply the same budgetization process used for other summaries.	

Our final prompt was:

\texttt{"For this meeting \{transcript\}, please extract the numbers corresponding to the important utterances for the creation of an extractive summary. Respond with only the number of the important utterances and nothing else."}

\subsection{Discourse parsing} \label{app:a2}

Discourse parsing has seen relative success through the application of large language models to frameworks like RST \citep{maekawa-etal-2024-obtain} and SDRT \citep{thompson2024llamipa}. Despite these advancements, using language models for discourse parsing remains relatively new and evolving. Existing parsers are trained on the STAC dataset, which lacks examples of corrections, parallels, and background, leading to bias when applied to AMI (see Figure \ref{fig:frequency}). Therefore, leveraging GPT-4 (which is not explicitly trained on STAC) with few-shot learning for discourse parsing can yield a more organic distribution of discourse relations. This approach is advantageous because it bypasses the constraints of pre-annotated corpora and offers a more "organic" or "unbiased" distribution of discourse relations, free from training bias.
The proposed parsing approach is straightforward. It utilizes the numbered utterances from extractive summarization and instructs the model to link them using one of the 16 SDRT relations, or "Other" if the relation isn't recognized as one of the predefined ones. The prompt provided to GPT-4 includes definitions of all 16 relations, along with examples of short conversations and their corresponding annotated graphs. Afterward, GPT-4 is given the meeting data and asked to generate a similar graph using the defined relations.

As shown in Figure \ref{fig:frequencyGPT}, certain relationships absent from STAC are more frequently predicted by GPT-4. For example, Corrections appear more prominently, while Contrasts, which are commonly predicted by deep sequential models, are notably underrepresented in GPT-4’s predictions.

Despite being relatively simple, this method shows great promise and warrants further investigation in future research. A major strength is its capacity to bypass training biases, reducing reliance on the limitations of small annotated datasets. Furthermore, unlike traditional parsing models that need specialized training for English text, GPT-4 is language-agnostic, offering increased flexibility across different linguistic contexts. As more models with stronger reasoning capabilities are developed, there is substantial potential for LLMs to greatly enhance discourse graph generation, further advancing discourse parsing techniques.

\section{Additional ablations studies}
\label{app:b}

\subsection{Relation type Ablation}

In Section \ref{sec:Abl}, we proposed examining the individual role of each discourse relation in the context of node classification. While this helps in understanding the isolated contribution of each relation, real-world meetings and decision-making processes rarely rely on just one. Instead, dialogues are driven by a combination of interconnected discourse functions. Ignoring these interactions by focusing on individual relations overlooks the synergistic dynamics at play.

In a scenario where a group discusses a potential solution to a problem, the conversation might start with a ``question-answer pair'' to mark the beginning of the decision-making process, followed by a ``result'' that signals the conclusion or decision reached by the group. However, focusing only on the ``result'' overlooks the subsequent ``acknowledgement'', which may indicate varying levels of agreement or dissent. Failing to maintain this synergy may result in hallucinations due to the omission of negation.

In this subsection, we propose to explore heuristic groupings of relations to investigate how they may work together to provide a more holistic understanding of dialogues. We propose to study the impact of four distinct groups of relations: ``Acknowledgement, Continuation, Elaboration'', which are the three most frequently occurring relationships; ``Continuation, Elaboration, Result'', which represent the most common relationships adjacent to extractive nodes; ``Question Answer Pair, Result, Acknowledgement'', which highlight decision-making processes; and ``Question Answer Pair, Explanation, Elaboration, Result'', a broader set of relations that captures question-driven interactions and explanatory connections, providing a comprehensive view of extractive summarization dynamics. This analysis will help us better understand how these relation groups influence the summarization process across different contexts.

\begin{figure}[t]
\centering
\includegraphics[width=\columnwidth]{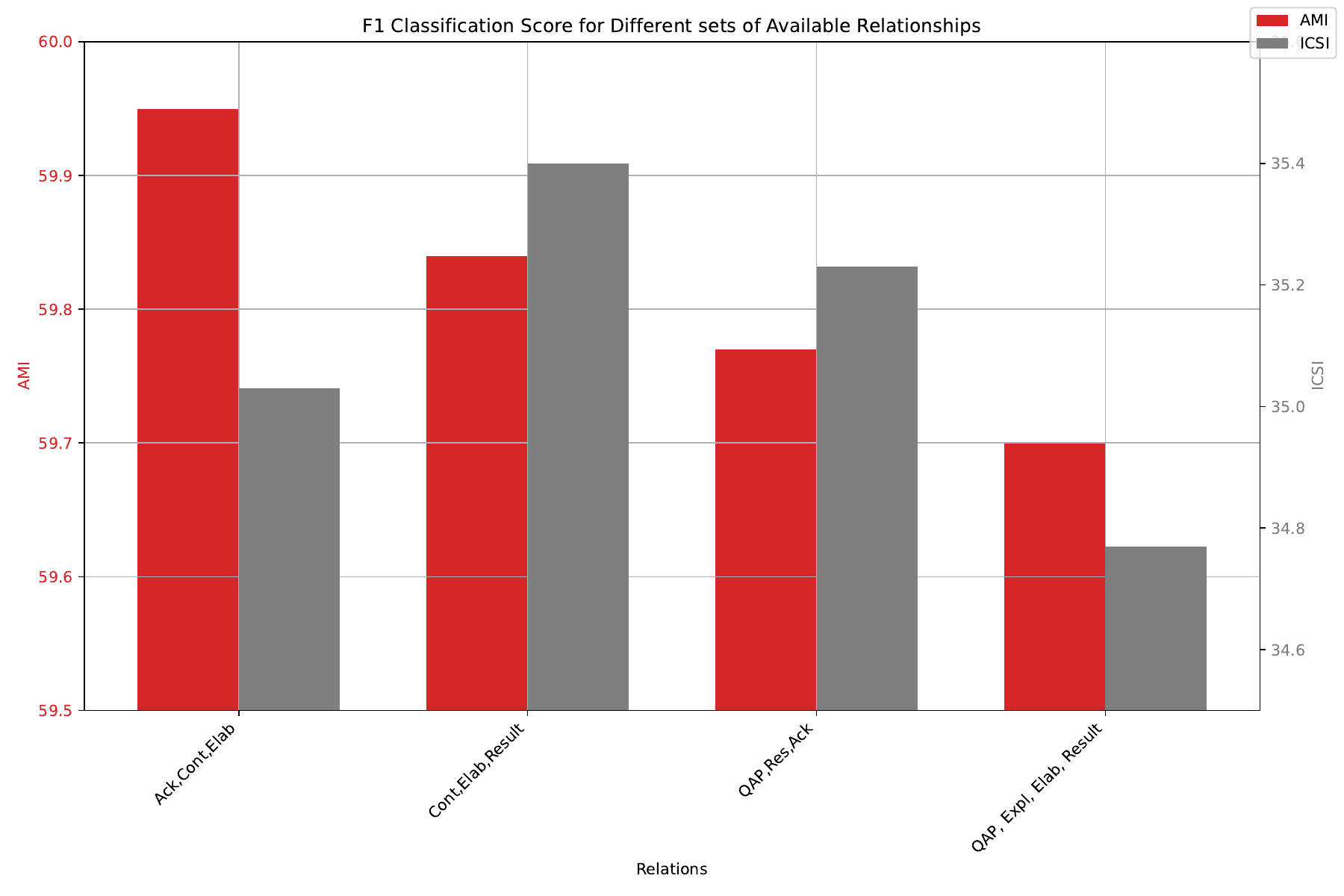}
\caption{Impact of the sets of discourse relations on our RGCN classifier for both AMI and ICSI}
\label{fig:Sets}
\end{figure}

When studying the rate at which the different relations appear and the synergy of different sets of relations; we can observe on Figure \ref{fig:Sets}, and compare these findings to Figure \ref{fig:ablation}. We observe that while different sets of relations have their own impact, they do not consistently outperform specific individual relations. This discrepancy may be attributed to several factors. One key reason is that combining multiple relations within a set can introduce overlapping or redundant information. In such cases, the model may extract superfluous details, focusing on peripheral aspects of a discussion rather than core decisions, which are often the most relevant for extractive summarization. Although theoretically, the inclusion of more relations should enhance understanding, the reality is more nuanced. The limited availability of meeting data for training and the potential noise introduced during graph generation may cause models to perform better when focusing on fewer, more distinct types of relations. For example, relations that directly indicate decisions, such as pointing to results in ICSI, may be easier for models to learn and leverage effectively. Moreover, the complexity of processing larger sets of relations could inadvertently obscure the most critical elements of the discussion. In such cases, concentrating on more targeted, easily learnable information, such as decision-related relations, may yield better performance by simplifying the task for the model and reducing noise. 

This underscores the importance of carefully selecting relations that provide the most relevant and distinct information, rather than assuming that larger sets of relations will always lead to improved results.

\end{document}